# Person Re-Identification without Identification via Event Anonymization

Shafiq Ahmad[1,2]   Pietro Morerio[1]   Alessio Del Bue[1]

[1]Pattern Analysis & Computer Vision (PAVIS) - Istituto Italiano di Tecnologia, Italy   [2]Universita degli Studi di Genova, Italy

{shafiq.ahmad, pietro.morerio, alessio.delbue}@iit.it

## Abstract

*Wide-scale use of visual surveillance in public spaces puts individual privacy at stake while increasing resource consumption (energy, bandwidth, and computation). Neuromorphic vision sensors (event-cameras) have been recently considered a valid solution to the privacy issue because they do not capture detailed RGB visual information of the subjects in the scene. However, recent deep learning architectures have been able to reconstruct images from event cameras with high fidelity, reintroducing a potential threat to privacy for event-based vision applications. In this paper, we aim to anonymize event-streams to protect the identity of human subjects against such image reconstruction attacks. To achieve this, we propose an end-to-end network architecture jointly optimized for the twofold objective of preserving privacy and performing a downstream task such as person ReId. Our network learns to scramble events, enforcing the degradation of images recovered from the privacy attacker. In this work, we also bring to the community the first ever event-based person ReId dataset gathered to evaluate the performance of our approach. We validate our approach with extensive experiments and report results on the synthetic event data simulated from the publicly available SoftBio dataset and our proposed Event-ReId dataset. The code is available at* https://github.com/IIT-PAVIS/ReId_without_Id

## 1. Introduction

For security and monitoring purposes, intelligent surveillance systems are installed in our personal spaces (e.g., home surveillance) and all over urban areas (hospitals, banks, shopping malls, airports and streets, etc.). However, collecting images and videos with always-connected vision sensors raises new issues: *i)* ethical discussions over the balance between safety/security needs and individual privacy; *ii)* unauthorized access to sensory data that may threaten users' privacy; *iii)* extensive resource consumption of large-scale sensor networks, e.g., energy, bandwidth, and computing power. Neuromorphic vision sensors (event cameras) are a disruptive technology as they only capture scene dynamics and do not record visual detail of humans, which enforces privacy-by-design (to some extent); their ultra-low resource consumption makes them ideal for always-on visual sensors. Besides, their high dynamic range enables them to work under challenging illumination conditions while, alike RGB cameras, event-cameras are able to solve various vision tasks, such as object recognition [26], human pose estimation [33], detection and tracking [27, 15, 21], and person re-identification (ReId) [1].

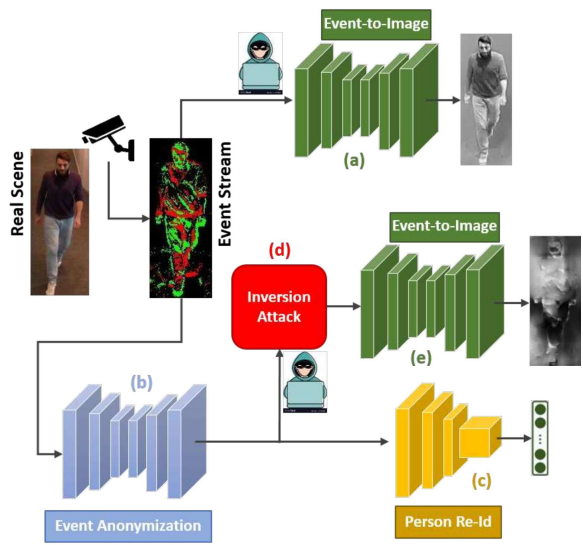

Figure 1. Event-to-image [32] can be regarded as a *privacy attack*, which reconstructs the appearance of a person from an event stream **(a)**. We propose a learnable Event Anonymization network architecture **(b)**, which deals with such attack by scrambling the event stream so that reconstruction deteriorates while preserving the performance of an event-based downstream task (e.g., person ReId **(c)**). We also consider a possible *Inversion Attack* **(d)**, where the attacker tries to reverse the effect of the proposed anonymization in order to attain image reconstruction **(e)**.

Event cameras output asynchronous events that are triggered with extremely low latency when an intensity change at pixel level is over a given threshold. Due to their asynchronous nature, event streams do not form images but rather a data-stream containing pixel position activation

(i.e., $(u, v)$ coordinates) and a polarity. These event streams were considered privacy-preserving, as they do not contain detailed visual features that can let a human or algorithm recognize individual traits such as faces. However, event streams encode the entire visual signal in an extremely compressed form and could, in principle, be decompressed to recover a high-quality video stream. Currently, deep neural network-based image reconstruction models [32, 30, 37, 43] have demonstrated impressive abilities in recovering grayscale images from event streams, representing a potential threat to the privacy of event-based vision applications as shown in Fig. 1 (a).

Recently, Du et al. [11] proposed a hand-crafted encryption framework to prevent privacy attacks on event-streams. Their approach incorporates a spatial chaotic mapping to scramble the positions of events and flip their polarities. The spatial information in the encrypted event-stream is thus deformed due to 2D position scrambling and, as a result, event-to-image methods fail to reconstruct high-quality images. The main drawback of this encryption technique is that downstream computer vision tasks cannot be performed directly with the encrypted event-stream, which is only useful to protect data during transmission or storage and must be decrypted before being utilized.

In this paper, we propose a learning-based approach called Event-Stream Anonymization which prevents privacy attacks on event data (see Fig. 1 b), while at the same time allowing the execution of downstream tasks. The proposed method enforces the degradation of images recovered from the privacy attacker (i.e., event-to-image module) while jointly optimizing a downstream task in an end-to-end fashion. In other words, it helps protect subjects' identity while preserving the information needed to achieve other tasks, such as person ReId as shown in Fig. 1 (c). The two tasks, anonymization and ReId, seem to have contrasting objectives. This represents the actual challenge of our work. However, ReId only aims at associating images of a person in a camera network, while anonymity refers to protecting a person's identity or other biometric traits. A practical use case is when an attacker has a person's name and photo and aims at identifying that person by maliciously accessing a camera network. The proposed anonymization pipeline prevents this attack while allowing ReId by the surveillance system.

We verify that our approach can successfully anonymize the event-stream with only a small drop in performance in person ReId by performing extensive experiments on simulated event data and on a newly introduced real event-based person ReId dataset called Event-ReId. More specifically, to evaluate the robustness of our method against event-to-image reconstruction techniques, not only do we measure the (poor) quality of the recovered images, but we also verify that classic full-body human identification or face identification tasks are hardly possible using such anonymized data. In addition, we validate the robustness of our anonymization technique against an inversion attack, where an attacker tries to reverse the effect of the anonymization network (see Fig. 1 d).

The main contributions of this work are summarised as follows:

- We propose an event-stream anonymization network to protect the identity information against event-to-image attacks in event-based vision applications. We also propose a joint optimization framework that preserves anonymization with a small drop in performance while testing on downstream tasks (e.g., ReId).
- We contribute a first-ever person ReId dataset captured with event camera, namely the Event-ReId dataset.
- We performed extensive experiments to verify the robustness of event stream anonymization network against privacy attacks (event-to-image approaches) using synthetic and the proposed real event dataset.

## 2. Related Work

### 2.1. Privacy-Preserving Computer Vision

**Standard (RGB) Vision Sensor:** Currently, few methods are developed to solve privacy-preserving issues for standard RGB cameras. These methods [18, 17, 20, 38, 36, 12, 7] can be divided into software and hardware level protection against privacy attacks. Methods based on software-level protection [20, 38, 12, 7] employ various computer vision algorithms to morph image/video data representations after acquisition. Thus, they learn privacy-preserving encodings through adversarial training to degrade privacy-related visual information in images/videos while trying to preserve essential features to perform inference tasks and prevent adversarial attacks. The hardware-level protection framework acts instead on the vision sensor to include an additional layer of security by removing sensitive data during the image acquisition. Most recent approaches optimize the distortion parameters of a virtual lens via adversarial training to hide the identity information of humans while allowing essential visual information to be gathered for computer vision task [18, 17, 36]. Actual lenses can then be manufactured using the learned coefficients.

**Event-based Vision Sensors:** Event cameras are often regarded as privacy-preserving as they naturally discard detailed visual biometric information (such as face details). However, the events-stream encodes the complete visual signal in an extremely compressed form and recent works were able to decompress it and recover a standard (grayscale) visual output, either using patch-based dictionaries [3], variational models [28], or deep learning-based solutions [32, 30, 37, 43]. Such event-to-image conversion approaches seem to suggest that event-cameras can no longer consider privacy-preserving devices since attackers can train their own models to break anonymity. Du et al. [11] investigate the privacy of event cameras and analyze the possible security attacks, including gray-scale image reconstruction and privacy-related classification. In addition, to prevent event-to-image conversion approaches, they proposed a hand-crafted encryption framework that incorporates spatial chaotic mapping to scramble the positions of events and polarity flipping. However, this framework is only useful to protect event-stream during transmission and storage purposes. In fact, the visual information is distorted in the event-stream due to 2D position scrambling and a computer vision module (e.g., person ReId, tracking, detection, etc.) can not directly be applied to the encrypted event-stream. On the contrary, we develop a method that distorts event-stream in such manner that image reconstruction methods produce degraded images while preserving the useful information to perform computer vision tasks (e.g., person ReId) on those distorted events.

## 2.2. Person Re-Identification

Person Re-Identification has gained significant interest as an enabling technique for smart video surveillance systems (e.g., tracking in non-overlapping views, forensic and security applications [40]). The person ReId problem has been extensively studied in standard (RGB) camera networks and deep-learning-based ReId approaches [39, 40] have improved the performance rapidly. Most of the existing ReId frameworks are developed for conventional RGB cameras, although different methods have been proposed for multi-modal person ReId such as e.g., cross-modal RGB-infrared [6, 22] and with RGB-D camera [2, 29].

Nowadays, ReId raises severe privacy concerns and preserving people's privacy becomes essential [10] also in view of the General Data Protection Regulation (EU GDPR). Currently, very few methods [9, 41, 10] addressed privacy concerns in person ReId. Julia et al. [9] apply face blurring to anonymized person identity and perform ReId. On the other hand, Shuguang et al. [10] suggested a privacy-preserving ReId method called person identity shift (PIS) that removes the absolute identity of the image (i.e., who is the person in the image) while preserving the relationship between images pairs. Zhao et al. [41] proposed a cloud-based privacy-preserving solution for ReId. That allows the cloud server to perform person ReId operations on encrypted data and output the final ReId results in plain text.

A major drawback of all the above methods is that they did not ensure the end-to-end privacy of the ReId system. The possibility of unauthorized access to the surveillance camera still poses severe threats to privacy. To address this challenge, the authors in [1] proposed an event-based person ReId system. Since event cameras capture scene dynamics without providing RGB image content, Ahmad et al. [1] showed that event-frames deliver mostly edge and texture contours details that might be used for ReId. Nevertheless, as already discussed, event-based streams still disclose personal traits by using neural networks [32, 37, 43] that can extract high-quality grayscale images from event-stream. To achieve an end-to-end privacy-preserving person ReId system, we propose a learning-based approach called event-stream anonymization for privacy-preserving person ReId. Our model learns to anonymize the event-stream to prevent image reconstruction techniques (i.e., privacy attacks) from recovering gray-scale images that may disclose identity information.

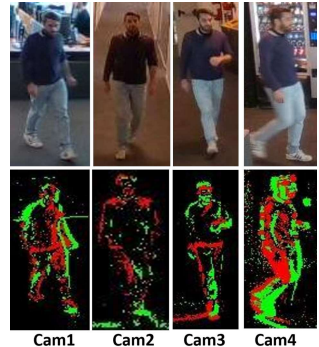

Figure 2. Samples from RGB and event cameras views from our Event-ReId dataset.

Table 1. Event-based dataset size comparison.

| Dataset | Event ReID | n-HAR | DailyAction DVS | DHP19 |
|---|---|---|---|---|
| No. of subjects | 33 | 30 | 15 | 17 |

## 3. Event-ReId: A New Dataset and Benchmark

Our target is to develop privacy-preserving person ReId methods using event-cameras. Yet, the research community lacks a dataset captured with real event cameras, which are also suitable for benchmarking person ReId methods. Hence, despite the advantages of event cameras in a surveillance application, research has been held back by the unavailability of event data, and so far, only simulated experiments have been deployed [1]. To address this issue and to boost new research on this topic, we propose the first event-based person ReId dataset named **Event-ReId**.

The Event-ReId dataset comprises 33 subjects walking across a non-overlapping field of view of four Prophesee® integrated within a surveillance network. The cameras feature different positioning and tilt angles; each one is coupled with an RGB camera in a fixed stereo configuration that captures approximately the same scene and being both synchronized by the network clock, see Fig. 2. Each RGB camera records data at 30 FPS at a resolution of $640{\times}480$ pixels, having captured a total of 16K images with an average of 120 frames per person per camera. The event-cameras resolution is the same as the RGB camera and each stream is recorded for the same length (≈4 sec) for both sensors. Further, out of 33 identities, 9 people wear face masks and each person appears in the view of all four camera pairs. The dataset includes variations, such as changes in illumination, pose, and viewpoint. We manually annotate the person and face bounding boxes on both event and RGB streams; the event ground truth bounding box is synchronized with RGB bounding boxes. The proposed dataset size compares favorably with the size of other event-based datasets: activity recognition dataset **n-HAR** [31] and **DailyAction-DVS** [23], and human pose estimation dataset **DHP19** [5] (see comparisons in Table 1).
Download **Event-ReId** from here https://doi.org/10.5281/zenodo.8256439

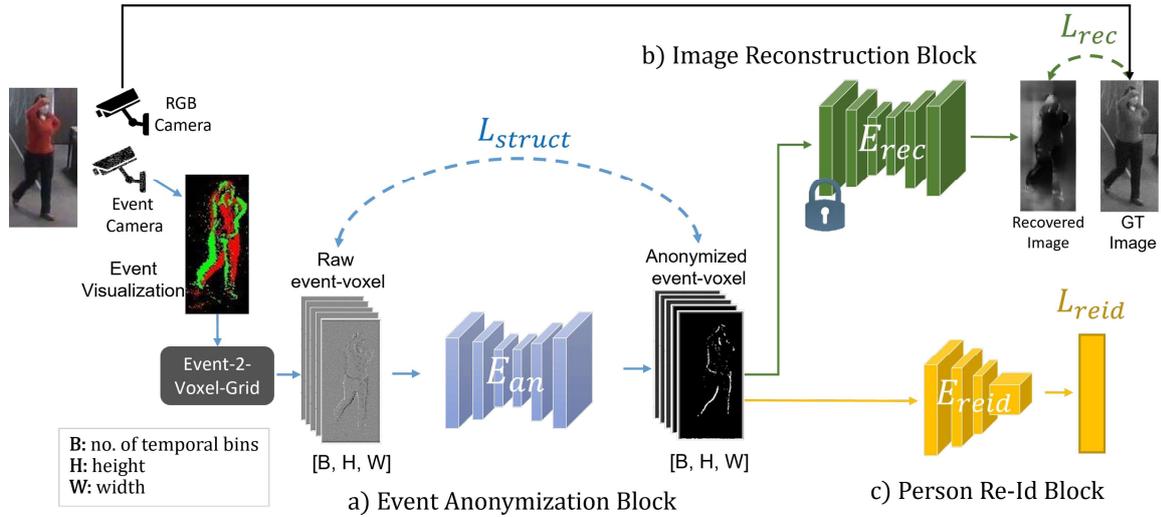

Figure 3. The complete pipeline of the proposed method. a) Event anonymization Network $E_{an}$ take raw event (voxel-grid) data and output anonymize event. $L_{struct}$ loss enforce $E_{an}$ to preserve the structural information in voxel-grid. b) Image reconstruction block $E_{rec}$ (pre-trained E2VID [32]) play the role of privacy attack and try to reconstruct grayscale image, $E_{an}$ maximize $L_{rec}$ loss to protect person identity information. c) Person ReId backbone $E_{reid}$ is trained with anonymized event data in an en-to-end fashion with $E_{an}$.

## 4. Proposed Method

The proposed pipeline consists of three main modules: the event anonymization block, the event-to-image reconstruction block, playing the role of the privacy attacker, and the person ReId block, which performs the downstream vision task on the anonymized event stream as shown in Fig 3. In the following, after describing the input event representation to the network, we provide a detailed description of each module, including its implementation and functionalities for preserving privacy and person ReId. We conclude this section with a description of the joint optimization method.

### 4.1. Input Event Representation

The output of an event camera is an asynchronous event stream that encodes the time, location, and polarity of the intensity changes (increase or decrease in intensity) [13]. Consequently, each event alone carries limited information about the scene appearance. Typically, asynchronous event data are converted to a grid-like representation as event-frame, or 2D histogram [25], time surface 2D map [34], and voxel grids [42]. This pre-processing facilitates the visualization and the extraction of meaningful information using standard frame-based methods such as deep convolutional neural networks (CNN) [25, 34, 42].

The input of our network is a voxel grid $X_e$ as proposed in [42]. A voxel grid is a space-time (3D) histogram of events generated by discretizing the time domain, where each voxel represents a particular pixel and time interval. Spatiotemporal coordinates, $x_k, y_k, t_k$, lie on a voxel grid such that $x_k \in \{1, 2, ..., W\}$, $y_k \in \{1, 2, ..., H\}$, and $t_k \in \{t_0, t_0 + \Delta t, ..., t_0 + B\Delta t\}$, where $t_0$ is the first time stamp, $\Delta t$ is the bin size, and $B$ is the number of temporal bins and W, H are the sensor width and height. We utilize a voxel grid representation for three reasons: *i)* to make the model fully differentiable; *ii)* the event-to-image methods in our proposed model also rely on a voxel grid; *iii)* a voxel grid preserves the temporal information of event streams.

### 4.2. Networks and modules

**Event-Stream anonymization Block.** The anonymization network in our framework (Fig. 3a) modifies the event-streams to prevent the following image reconstruction tech-niques from converting events into intensity images that can reveal privacy-sensitive information (e.g., faces). At the same time, this module should preserve useful spatial information needed for performing person ReId successfully. The anonymization network consist in a convolutional autoencoder [35] $E_{an}$ which takes a raw event-voxel $X_e \in \mathbb{R}^{B \times W \times H}$ and output anonymized event-voxel $\hat{X}_e \in \mathbb{R}^{B \times W \times H}$. The use of an autoencoder-like architecture is primarily justified by the fact that this module, in the worst-case scenario, should be able to replicate the event-stream in order to allow performing the downstream task. The autoencoder architecture consists of 4 convolutional layers, each with a filter size of 3 and a stride of 1.

**Image Reconstruction Block.** The image reconstruction module consists of a pre-trained E2VID network [32] that is a recurrent neural network that reconstructs high-quality grayscale images from the stream of events. In this block, any event-to-image method e.g., [30, 37, 43] can be integrated as a privacy attacker. E2VID translates a continuous stream of events into a sequence of images. To achieve this, the incoming stream of events is partitioned into sequential (non-overlapping) spatiotemporal windows. Similarly, we partitioned the input event streams into a fixed time window ***T*** (explained in Sec. 4.1) for the anonymization network $E_{an}$. The output voxel-grid $\hat{X}_e$ is then processed by reconstruction module $E_{rec}$ to reconstruct the target grayscale image. We thus encourage degradation in the recovered image to prevent identity information leakage. Note that the weights of this module are not updated during training.

**Event-based Person ReId Block.** Person ReId methods aim to learn a vector representation, usually a feature embedding from a CNN, of images to perform retrieval and recover images belonging to the same person Id. In our case, ReId is performed on event-stream data instead of the standard RGB signals.

We employ a ResNet-50 [16] pre-trained on ImageNet as the backbone for feature embedding. Unlike the event-based ReId in [1], which utilizes event-frames, our ReId module $E_{reid}$ takes anonymized event-voxels $\hat{X}_e \in \mathbb{R}^{B \times W \times H}$ as input. We modify the original ResNet architecture to accommodate the $B$ input channels of the voxel-grid representation and compute a 256-D feature embedding for ReId. The ReId model uses classification loss (cross-entropy) and triplet loss for all experiments and is jointly trained with the anonymization network.

### 4.3. End-to-End Training

Our ultimate goal is to learn the parameters of anonymization network $E_{an}$ such that: *i)* event-to-image techniques cannot recover intensity image from $E_{an}$ output that can disclose private visual information; *ii)* person ReId achieves the best performance or at least does not experience a significant drop if compared to using a non-anonymized event-stream. The three modules are combined as shown in Fig. 3 so that the output of $E_{an}$ (anonymized stream) is the input of $E_{rec}$ and $E_{reid}$ at once. We train all the modules jointly in an end-to-end manner, described in detail below. $E_{an}$ has the aim of neutralizing the reconstruction attack, thus ultimately must be trained with the objective of degrading the quality of the recovered images $\hat{I}_{image} = E_{rec}(E_{an}(X_e))$. To this end, we use the structural similarity index (SSIM) [19] to assess the quality of $\hat{I}_{image}$ compared to the ground-truth $I_{image}$:

$$\mathcal{L}_{rec} = SSIM(\hat{I}_{image}, I_{image}). \quad (1)$$

SSIM is one of the most popular perception-based error metrics [35], aiming to measure better image luminance, contrast, and structure information. Since our objective is to degrade the recovered image, the SSIM function is bounded ranges between [0-1], where a value near 0 indicates less similarity between two compared images. Thereby the $\mathcal{L}_{rec}$ loss is minimized during training to force the images recovered by the attacker to be as more diverse as possible from the real ones. Moreover, as our anonymization model scrambles the input raw event-voxel, the useful visual information in the event-voxel could be lost, decreasing the performance of the person ReId substantially. To preserve the structural similarity between $X_e$ and $\hat{X}_e$, which is useful for person ReId, we compute the structural loss as

$$\mathcal{L}_{struct} = 1 - SSIM(\hat{X}_e, X_e), \quad (2)$$

and define the person ReId objective as

$$\mathcal{L}_{reid} = \mathcal{H}(P_{id}, E_{reid}(\hat{X}_e)). \quad (3)$$

Here $\mathcal{H}$ refers to the identity loss (cross entropy and triplet loss) function and $P_{id}$ is the ids label for person ReId. Thus, our training scheme jointly models the event-stream anonymization with person ReId during training and the overall cost function can be written as:

$$\mathcal{L}_{Total} = \alpha \mathcal{L}_{struct} + \beta \mathcal{L}_{rec} + \gamma \mathcal{L}_{reid}. \quad (4)$$

## 5. Experiments

**Synthetic and real datasets** We test our method on reconstruction and inversion attacks using synthetic data and the real dataset presented in Section 3. Synthetic event data is generated from the video-based person ReId Soft-Bio [4] dataset through open-source event simulator [14]. The SoftBio dataset comprises 152 identities and a total of 64,472 frames collected with eight surveillance cameras. The dataset is recorded in an uncontrolled environment, and each identity may only appear in a subset of cameras, which collect data under very different viewpoints, with drastic changes in illumination and background. In addition, we benchmark our approach on the **Event ReId** dataset described in Sec. 3.

**Setup and Implementation:** We simulate event data from SoftBio, randomly splitting 152 identities, 76 IDs for training and 76 other IDs for testing. For the real data in Event ReId, we randomly select 22 IDs for training and 11 IDs for testing out of 33 identities of the proposed real event-based person Event-ReId.

We choose the time span for the spatiotemporal voxel grid $T \approx 40ms$ for synthetic event data and $T \approx 33.3ms$ for real event data to be synchronized with the corresponding RGB frames. Following [32], we set the size of temporal bin $B = 5$ for the event voxel grid and during training, our model resized the event voxel grid to $5 \times 392 \times 192$. We use a batch size of 24 and train the model with a base learning rate of 0.001 for 60 epochs. We set momentum $\mu = 0.9$ and the weight decay to $5 \times 10^{-4}$. In Eq. 4 we set $\alpha = \beta = \gamma = 1$. The implementation is based on PyTorch.

### 5.1. Metrics and Evaluation Methods

To evaluate the performance of our complete model on reconstruction and inversion attacks, we need to assess the trade-off between person ReId and privacy-preserving tasks. We first processed the raw event-stream for both tasks during inference through our anonymization network to acquire the anonymized event data. Later, we measure the performance of person ReId and privacy-preserving tasks using anonymized data.

**Person ReId:** Our main goal is to perform person ReId with anonymized event data without compromising ReId accuracy. We thus train our ReId backbone on both anonymized and raw events separately and then compare their performance. We report the rank accuracy and mean average precision for both real and simulated data.

**Privacy-preserving:** Here, we consider the case in which the attacker has access to the anonymized event data and tries to disclose the person's identity by employing image reconstruction, e.g., E2VID [32]. To experimentally test the robustness of our event stream anonymization approach against the reconstruction attack, we measure the image quality using the structural similarity index (SSIM) and peak-signal-to-noise ratio (PSNR). Low values of SSIM and PSNR suggest low image quality, which is what we expect to achieve if anonymization is successful. We compute the average SSIM and PSRN for all images in test sets of the real and simulated datasets.

In addition, we also validate that our proposed identity anonymization framework completely removes information that can be used to identify the persons. Therefore, we also formulate the privacy attack as an image retrieval and face verification task.

*(i) Image Retrieval:* We consider that an attacker has access to the event-based privacy-preserving surveillance

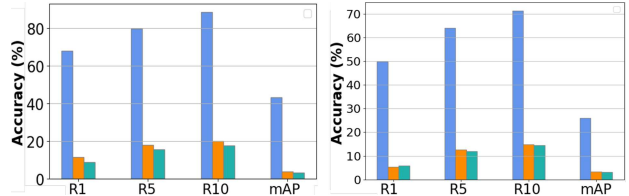

Figure 4. Image retrieval score on Event-ReId (left) and SoftBio (right), following query-gallery setting, blue: $Q_{RGB}$, $G_{event}$, orange: $Q_{event}$, $G_{an-event}$ and green: $Q_{RGB}$, $G_{an-event}$

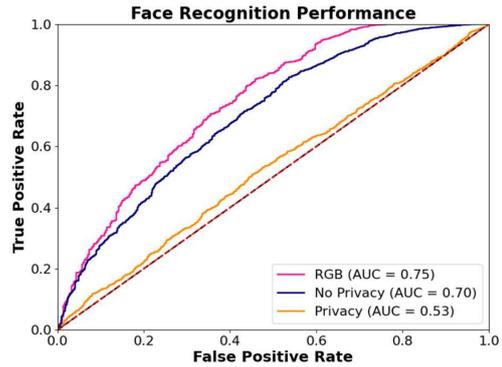

Figure 5. Face recognition accuracy using Arcface[8] model.

Table 2. Recovered image quality: SSIM and PSNR values.

| Dataset | **Real** | | **Synthetic** | |
|---|---|---|---|---|
| **Method** | SSIM↓ | PSNR↓ | SSIM↓ | PSNR↓ |
| No-privacy | 0.548 | 11.617 | 0.530 | 11.284 |
| Privacy (Our) | 0.384 | 8.943 | 0.368 | 8.071 |

camera network and also holds a query image of a target to identify. The query image is either captured with a standard RGB camera $Q_{rgb}$ or a gray-scale image $Q_{event}$ reconstructed from an event-stream without the protection of the privacy module. Then the attacker determines whether this person exists in the gallery set $G_{an-event}$ that contains degraded images by using the query image to retrieve the correct target identity. Higher retrieval performance indicates a lower privacy-preserving effect: $E_{an}$ performance is evaluated based on the rank accuracy or mean average precision metrics. For this experiment, we employ the state-of-the-art person ReId model BOT [24] to evaluate image retrieval and use the test sets of real and simulated datasets.

*(ii) Face Recognition:* In this experiment, we assume a similar scenario, where the attacker holds a *face* image (RGB or reconstructed gray-scale image) and tries to disclose identity information by matching it with a degraded face image. We use the pre-trained face recognition model ArcFace [8] to measure the resilience of our system to this privacy attack. We measure face recognition performance in terms of the area under the curve (AUC) of the ROC curve.

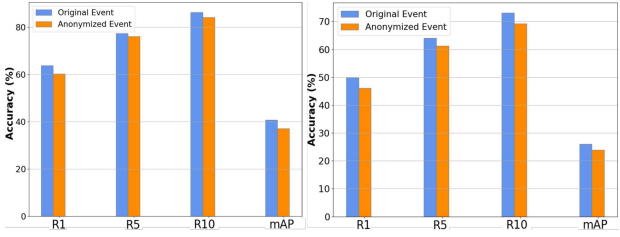

Figure 6. Person ReId performance on Event ReId dataset (left) and SoftBio dataset (right) using raw and anonymized event.

Table 3. Comparison with other methods on the Event-ReId dataset: image quality, event-reid, and image retrieval.

| Method | SSIM↓ | PSNR↓ | $R1_{reid}$↑ | $R1_{retr}$↓ |
|---|---|---|---|---|
| No-privacy (raw events) | 0.548 | 11.617 | 63.7 | 67.8 |
| Encryption$_{Discarding}$ [11] | 0.486 | 10.57 | 38.2 | 41.1 |
| Encryption$_{Scrambling}$ [11] | 0.451 | 10.12 | 29.4 | 30.9 |
| **Privacy (Ours)** | **0.384** | **8.943** | **59.2** | **8.9** |

Table 4. Person Re-Id Performance on Event-ReId dataset.

| Method | Privacy | R1 | R5 | R10 | mAP |
|---|---|---|---|---|---|
| ED-ReId[1] | No | 57.9 | 75.8 | 82.3 | 34.9 |
| Ours | No | 63.7 | 77.3 | 86.2 | 40.7 |
| Ours | Yes | 59.2 | 76.1 | 84.1 | 36.1 |

## 5.2. Results on Reconstruction attack

**Privacy-preserving performance.** We present the image retrieval performance score for the real dataset Event-ReId and similarly for the simulated event data of SoftBio in Fig. 4. The testing approaches "$Q_{rgb} \Leftrightarrow G_{an-event}$" and "$Q_{event} \Leftrightarrow G_{an-event}$" measure the retrieval score on the anonymized (privacy-preserving) image gallery using original RGB and recovered gray-scale query images respectively. For comparison, the testing approach "$Q_{rgb} \Leftrightarrow G_{event}$" measures the image retrieval score on original gallery images. The tested retrieval model BoT [24] did not perform well on our anonymized images and for both datasets the retrieval score is random.

Regarding the face recognition performance, Fig. 5 shows the ROC curves for each testing approach: **RGB** measures the face verification score on original RGB face images; **No Privacy** measure face verification score between RGB and gray-scale face images recovered from original events; **Privacy** measure face verification score between RGB and gray-scale images recovered from anonymized events. From the figure, we can conclude that the ArcFace model does not perform well on the images reconstructed from the anonymized event-stream as the AUC= 0.53, which is close to the random performance.

In addition, Table 2 presents image quality measurement using SSIM and PSNR. As observed, lower values for SSIM and PSNR are associated with degraded images. Results from all three experiments: image retrieval, face recognition, and image quality assessment, suggest that our pro-posed event anonymization network successfully preserves the person's identity information.

**Person ReId Performance.** The rank accuracy and mean average precision score of person ReId utilizing anonymized event data for both real Event-ReId and SoftBio[4] datasets are presented in Fig. 6. The results show that the anonymization model does not affect the person ReId performance. While shifting from *No Privacy* to *Privacy-preserving*, the R1 accuracy and mAP drop is 3.5% and 3.6%, respectively, for Event-ReId data. Similarly, for SoftBio, the drop in R1 is 3.8% and for mAP is 2.1%.

**Comparison with baselines.** Since this work investigates privacy-preserving person ReId for the first time, there is no other method for direct comparison. We benchmark our approach against event encryption (partial scrambling and discarding) methods [11] to check their effect on privacy-preserving. We use partial (75%) encryption for both the scrambling and discarding algorithms, as complete encryption distorts the entire visual information in the event data, which can not be utilized for downstream tasks. Table 3 reports SSIM and PSNR image quality metrics, $R1_{reid}$: Rank1 accuracy of person ReId, and $R1_{retr}$: Rank1 accuracy of image retrieval on reconstructed images, using proposed Event ReId dataset. Our proposed event anonymization method outperforms the event encryption technique.

Further, we compare the person ReId, with a baseline event-driven ReId method (Ed-ReId) [1]. Results in Table 4 illustrate that even after event-stream anonymization with our proposed network, ReId performance is still better than Ed-ReId, although we pay a reasonable decrease in the score when applying privacy (i.e., anonymization module).

Table 5. Ablation on the losses for person ReId accuracy.

| $\alpha\mathcal{L}_{struct} + \beta\mathcal{L}_{rec} + \gamma\mathcal{L}_{reid}$ | Method | R1 | R5 | R10 |
|---|---|---|---|---|
| $\alpha = \beta = 0, \gamma = 1$ | Raw$_{Ev}$ | 63.7 | 77.3 | 86.2 |
| $\alpha = 0, \beta = \gamma = 1$ | anonymized$_{Ev}$ | 54.5 | 72.7 | 77.4 |
| $\alpha = \beta = \gamma = 1$ | anonymized$_{Ev}$ | 59.2 | 76.1 | 84.1 |

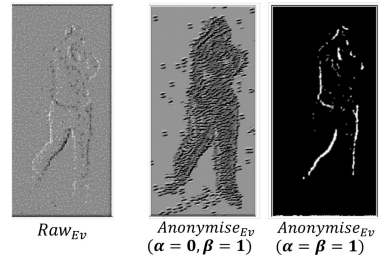

Figure 7. Raw event-voxel (left), anonymized event-voxel without $L_{struct}$ loss (middle) and with $L_{struct}$ loss (right).

**Losses ablation.** We finally analyze the losses' effect on the downstream task's accuracy (Person ReId). Without privacy ($\alpha=\beta=0$) the Rank1 accuracy is 63.7%, with privacy ($\alpha=0$, $\beta=1$) Rank1 accuracy significantly decreased to 54.5%. Finally, including $L_{struct}$ loss ($\alpha=\beta=1$) (helps to maintain structural information while anonymizing the voxel-grid) recovers the accuracy to 59.2%, still preserving privacy, as detailed in Table 5 and Fig 7.

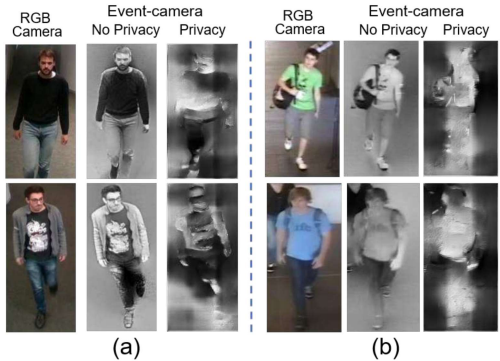

Figure 8. Visualisation of reconstructed images obtained using learning-based event anonymization method. a) real event dataset Event-ReId; b) synthetic event dataset SoftBio.

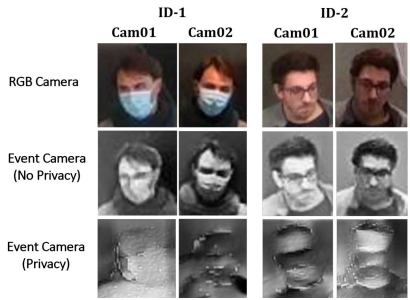

Figure 9. Visualisation of reconstructed face images. Top row RGB images. Middle row recovered from raw event. Bottom row recovered from anonymized events.

Table 6. Image retrieval performance on the Event-ReId dataset for recovered image under Inversion Attack.

| Method | R1 | R5 | R10 | mAP |
|---|---|---|---|---|
| No-Privacy | 67.8 | 79.9 | 88.4 | 40.7 |
| Privacy | 8.9 | 15.6 | 17.7 | 3.2 |
| Inversion Attack | 9.1 | 14.3 | 17.4 | 2.9 |

**Qualitative Results.** We qualitatively compare the reconstructed images acquired using our approach with the original images. We show the results on two examples from each Event-ReId and Softbio data video from the dataset. Fig. 8 displays anonymized images compared to the original RGB and recovered gray-scale images for reference. As observed, the image reconstructed from anonymized events degraded as compared to non-privacy images. We also show the two exemplar face reconstructions from real event data in Figure 9, showing that the subject face can not be reconstructed from our anonymized event stream compared to face reconstruction from the non-privacy event stream.

### 5.3. Results on Inversion attack

We explore a scenario where an attacker has access to our privacy-preserving event camera; they can produce a large training set containing anonymized event data along with their corresponding original event data. In such a case, the attacker can possibly train a network $E_{inv}$ trying to reverse the effect of $E_{an}$, leading to the reconstruction of high-quality grayscale images. To validate the robustness of our proposed framework to such privacy attacks, we train an autoencoder network on the real event dataset, similar to the $E_{an}$ network. The network takes as input the anonymized event stream from the pre-trained $E_{an}$ network and is trained to minimize the image reconstruction loss (instead of maximizing it).

Quantitative results in Table 6 show the performance score of image retrieval on reconstructed images, suggesting that reconstruction is significantly poor and identity information is still preserved. Fig. 10 presents qualitative results on two sample images, which show the image reconstruction failed to recover images correctly. Hence, the inversion attack could not reverse the effect of the event anonymization network.

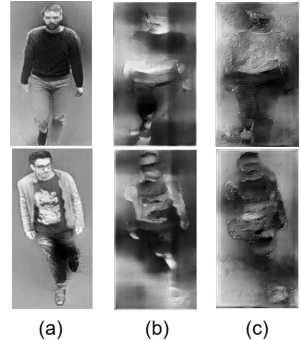

Figure 10. Reconstructed images from a) raw events, b) anonymized events, and c) output of Inversion Attack.

### 6. Conclusion

This paper presented an end-to-end learning-based approach for privacy-preserving person re-identification. We identified event-to-image techniques as a potential threat to privacy in event-based vision. The proposed approach jointly optimizes the event-stream anonymization to prevent privacy attacks while effectively performing ReId task. The proposed model is trained and evaluated on simulated event data and real event data Event-ReId. Human identification and face recognition results verify the efficacy of our framework against possible privacy attacks. We also demonstrate that our model is resistant to an inversion attack, which tries to reverse the effect of the anonymization module. The main limitations of the proposed pipeline are limited to a small drop in performance for the downstream task and a slight computational overhead due to the event anonymization network.

**Impact:** The proposed approach can be integrated with person ReId systems where privacy-preserving is essential. As this work aims to perform ReId tasks without disclosing human identity information, we believe that in the future, the event-stream anonymization mechanism can be extended to other event-based computer vision tasks to protect privacy at large. The potential negative impact lies in that surveillance data and person ReId datasets may be targeted by privacy attacks, which is why their acquisition, data storage, and protection should be strictly regulated. Also, the misuse of ReId can potentially have a negative impact.

## Acknowledgment

This work was partially supported by the project "RAISE-Robotics and AI for Socio-economic Empowerment" and has been supported by European Union-NextGenerationEU.

## References

[1] Shafiq Ahmad, Gianluca Scarpellini, Pietro Morerio, and Alessio Del Bue. Event-driven re-id: A new benchmark and method towards privacy-preserving person re-identification. In *Proceedings of the IEEE/CVF Winter Conference on Applications of Computer Vision (WACV) Workshops*, pages 459–468, 2022. 1, 3, 5, 7

[2] Igor Barros Barbosa, Marco Cristani, Alessio Del Bue, Loris Bazzani, and Vittorio Murino. Re-identification with rgb-d sensors. In *European Conference on Computer Vision*, pages 433–442. Springer, 2012. 3

[3] Souptik Barua, Yoshitaka Miyatani, and Ashok Veeraraghavan. Direct face detection and video reconstruction from event cameras. In *2016 IEEE winter conference on applications of computer vision (WACV)*, pages 1–9. IEEE, 2016. 2

[4] Alina Bialkowski, Simon Denman, Sridha Sridharan, Clinton Fookes, and Patrick Lucey. A database for person re-identification in multi-camera surveillance networks. In *2012 International Conference on Digital Image Computing Techniques and Applications (DICTA)*, pages 1–8. IEEE, 2012. 5, 7

[5] Enrico Calabrese, Gemma Taverni, Christopher Awai Easthope, Sophie Skriabine, Federico Corradi, Luca Longinotti, Kynan Eng, and Tobi Delbruck. Dhp19: Dynamic vision sensor 3d human pose dataset. In *Proceedings of the IEEE/CVF conference on computer vision and pattern recognition workshops*, pages 0–0, 2019. 4

[6] S Choi, S Lee, Y Kim, T Kim, and C Hi-CMD Kim. Hierarchical cross-modality disentanglement for visible-infrared person re-identification. In *Proceedings of the IEEE Conference on Computer Vision and Pattern Recognition, Seattle, WA, USA*, pages 13–19, 2020. 3

[7] Ishan Rajendrakumar Dave, Chen Chen, and Mubarak Shah. Spact: Self-supervised privacy preservation for action recognition. In *Proceedings of the IEEE/CVF Conference on Computer Vision and Pattern Recognition*, 2022. 2

[8] Jiankang Deng, Jia Guo, Niannan Xue, and Stefanos Zafeiriou. Arcface: Additive angular margin loss for deep face recognition. In *Proceedings of the IEEE/CVF conference on computer vision and pattern recognition*, pages 4690–4699, 2019. 6

[9] Julia Dietlmeier, Joseph Antony, Kevin McGuinness, and Noel E O'Connor. How important are faces for person re-identification? In *2020 25th International Conference on Pattern Recognition (ICPR)*, pages 6912–6919. IEEE, 2021. 3

[10] Shuguang Dou, Xinyang Jiang, Qingsong Zhao, Dongsheng Li, and Cairong Zhao. Towards privacy-preserving person re-identification via person identify shift. *arXiv preprint arXiv:2207.07311*, 2022. 3

[11] Bowen Du, Weiqi Li, Zeju Wang, Manxin Xu, Tianchen Gao, Jiajie Li, and Hongkai Wen. Event encryption for neuromorphic vision sensors: Framework, algorithm, and evaluation. *Sensors*, 21(13):4320, 2021. 2, 3, 7

[12] Joseph Fioresi, Ishan Dave, and Mubarak Shah. Tedspad: Temporal distinctiveness for self-supervised privacy-preservation for video anomaly detection. In *Proceedings of the IEEE/CVF International Conference on Computer Vision (ICCV)*, October 2023. 2

[13] Guillermo Gallego, Tobi Delbrück, Garrick Orchard, Chiara Bartolozzi, Brian Taba, Andrea Censi, Stefan Leutenegger, Andrew J Davison, Jörg Conradt, Kostas Daniilidis, et al. Event-based vision: A survey. *IEEE transactions on pattern analysis and machine intelligence*, 44(1):154–180, 2020. 4

[14] Daniel Gehrig, Mathias Gehrig, Javier Hidalgo-Carrió, and Davide Scaramuzza. Video to events: Recycling video datasets for event cameras. In *IEEE Conf. Comput. Vis. Pattern Recog. (CVPR)*, June 2020. 5

[15] Daniel Gehrig, Henri Rebecq, Guillermo Gallego, and Davide Scaramuzza. Eklt: Asynchronous photometric feature tracking using events and frames. *International Journal of Computer Vision*, 128(3):601–618, 2020. 1

[16] Kaiming He, Xiangyu Zhang, Shaoqing Ren, and Jian Sun. Deep residual learning for image recognition. In *Proceedings of the IEEE conference on computer vision and pattern recognition*, pages 770–778, 2016. 5

[17] Carlos Hinojosa, Miguel Marquez, Henry Arguello, Ehsan Adeli, Li Fei-Fei, and Juan Carlos Niebles. Privhar: Recognizing human actions from privacy-preserving lens. *arXiv preprint arXiv:2206.03891*, 2022. 2

[18] Carlos Hinojosa, Juan Carlos Niebles, and Henry Arguello. Learning privacy-preserving optics for human pose estimation. In *Proceedings of the IEEE/CVF International Conference on Computer Vision*, pages 2573–2582, 2021. 2

[19] Alain Hore and Djemel Ziou. Image quality metrics: Psnr vs. ssim. In *2010 20th international conference on pattern recognition*, pages 2366–2369. IEEE, 2010. 5

[20] Sudhakar Kumawat and Hajime Nagahara. Privacy-preserving action recognition via motion difference quantization. *arXiv preprint arXiv:2208.02459*, 2022. 2

[21] Gregor Lenz, Sio-Hoi Ieng, and Ryad Benosman. Event-based face detection and tracking using the dynamics of eye blinks. *Frontiers in Neuroscience*, page 587, 2020. 1

[22] Diangang Li, Xing Wei, Xiaopeng Hong, and Yihong Gong. Infrared-visible cross-modal person re-identification with an x modality. In *Proceedings of the AAAI Conference on Artificial Intelligence*, volume 34, pages 4610–4617, 2020. 3

[23] Qianhui Liu, Dong Xing, Huajin Tang, De Ma, and Gang Pan. Event-based action recognition using motion information and spiking neural networks. In *IJCAI*, pages 1743–1749, 2021. 4

[24] Hao Luo, Youzhi Gu, Xingyu Liao, Shenqi Lai, and Wei Jiang. Bag of tricks and a strong baseline for deep person re-identification. In *Proceedings of the IEEE/CVF conference on computer vision and pattern recognition workshops*, pages 0–0, 2019. 6, 7

[25] Ana I Maqueda, Antonio Loquercio, Guillermo Gallego, Narciso García, and Davide Scaramuzza. Event-based vision meets deep learning on steering prediction for self-driving cars. In *Proceedings of the IEEE conference on computer vision and pattern recognition*, pages 5419–5427, 2018. 4

[26] Nico Messikommer, Daniel Gehrig, Mathias Gehrig, and Davide Scaramuzza. Bridging the gap between events and frames through unsupervised domain adaptation. *IEEE Robotics and Automation Letters*, 7(2):3515–3522, 2022. 1

[27] Anton Mitrokhin, Cornelia Fermüller, Chethan Parameshwara, and Yiannis Aloimonos. Event-based moving object detection and tracking. In *2018 IEEE/RSJ International Conference on Intelligent Robots and Systems (IROS)*, pages 1–9. IEEE, 2018. 1

[28] Gottfried Munda, Christian Reinbacher, and Thomas Pock. Real-time intensity-image reconstruction for event cameras using manifold regularisation. *International Journal of Computer Vision*, 126(12):1381–1393, 2018. 2

[29] Marina Paolanti, Luca Romeo, Daniele Liciotti, Rocco Pietrini, Annalisa Cenci, Emanuele Frontoni, and Primo Zingaretti. Person re-identification with rgb-d camera in top-view configuration through multiple nearest neighbor classifiers and neighborhood component features selection. *Sensors*, 18(10):3471, 2018. 3

[30] Federico Paredes-Vallés and Guido CHE de Croon. Back to event basics: Self-supervised learning of image reconstruction for event cameras via photometric constancy. In *Proceedings of the IEEE/CVF Conference on Computer Vision and Pattern Recognition*, pages 3446–3455, 2021. 2, 5

[31] Bibrat Ranjan Pradhan, Yeshwanth Bethi, Sathyaprakash Narayanan, Anirban Chakraborty, and Chetan Singh Thakur. N-har: A neuromorphic event-based human activity recognition system using memory surfaces. In *2019 IEEE International Symposium on Circuits and Systems (ISCAS)*, 2019. 4

[32] Henri Rebecq, René Ranftl, Vladlen Koltun, and Davide Scaramuzza. High speed and high dynamic range video with an event camera. *IEEE transactions on pattern analysis and machine intelligence*, 43(6):1964–1980, 2019. 1, 2, 3, 4, 5, 6

[33] Gianluca Scarpellini, Pietro Morerio, and Alessio Del Bue. Lifting monocular events to 3d human poses. In *Proceedings of the IEEE/CVF Conference on Computer Vision and Pattern Recognition*, pages 1358–1368, 2021. 1

[34] Amos Sironi, Manuele Brambilla, Nicolas Bourdis, Xavier Lagorce, and Ryad Benosman. Hats: Histograms of averaged time surfaces for robust event-based object classification. In *Proceedings of the IEEE Conference on Computer Vision and Pattern Recognition*, pages 1731–1740, 2018. 4

[35] Jake Snell, Karl Ridgeway, Renjie Liao, Brett D Roads, Michael C Mozer, and Richard S Zemel. Learning to generate images with perceptual similarity metrics. In *2017 IEEE International Conference on Image Processing (ICIP)*, pages 4277–4281. IEEE, 2017. 5

[36] Zaid Tasneem, Giovanni Milione, Yi-Hsuan Tsai, Xiang Yu, Ashok Veeraraghavan, Manmohan Chandraker, and Francesco Pittaluga. Learning phase mask for privacy-preserving passive depth estimation. 2022. 2

[37] Wenming Weng, Yueyi Zhang, and Zhiwei Xiong. Event-based video reconstruction using transformer. In *Proceedings of the IEEE/CVF International Conference on Computer Vision*, pages 2563–2572, 2021. 2, 3, 5

[38] Zhenyu Wu, Haotao Wang, Zhaowen Wang, Hailin Jin, and Zhangyang Wang. Privacy-preserving deep action recognition: An adversarial learning framework and a new dataset. *IEEE Transactions on Pattern Analysis and Machine Intelligence*, 2020. 2

[39] Zizheng Yang, Xin Jin, Kecheng Zheng, and Feng Zhao. Unleashing potential of unsupervised pre-training with intra-identity regularization for person re-identification. In *Proceedings of the IEEE/CVF Conference on Computer Vision and Pattern Recognition*, pages 14298–14307, 2022. 3

[40] Mang Ye, Jianbing Shen, Gaojie Lin, Tao Xiang, Ling Shao, and Steven CH Hoi. Deep learning for person re-identification: A survey and outlook. *IEEE transactions on pattern analysis and machine intelligence*, 44(6):2872–2893, 2021. 3

[41] Bowen Zhao, Yingjiu Li, Ximeng Liu, Hwee Hua Pang, and Robert H Deng. Freed: An efficient privacy-preserving solution for person re-identification. In *2022 IEEE Conference on Dependable and Secure Computing (DSC)*, pages 1–8. IEEE, 2022. 3

[42] Alex Zihao Zhu, Liangzhe Yuan, Kenneth Chaney, and Kostas Daniilidis. Unsupervised event-based learning of optical flow, depth, and egomotion. In *Proceedings of the IEEE/CVF Conference on Computer Vision and Pattern Recognition*, pages 989–997, 2019. 4

[43] Lin Zhu, Xiao Wang, Yi Chang, Jianing Li, Tiejun Huang, and Yonghong Tian. Event-based video reconstruction via potential-assisted spiking neural network. In *Proceedings of the IEEE/CVF Conference on Computer Vision and Pattern Recognition*, pages 3594–3604, 2022. 2, 3, 5